\documentclass[letterpaper, 10 pt, conference]{ieeeconf}
\IEEEoverridecommandlockouts
\usepackage{amsmath,amsfonts}
\usepackage{algorithmic}
\usepackage{algorithm}
\usepackage{array}
\usepackage[caption=false,font=normalsize,labelfont=sf,textfont=sf]{subfig}
\usepackage{textcomp}
\usepackage{stfloats}
\usepackage{url}
\usepackage{verbatim}
\usepackage{graphicx}
\usepackage{cite}
\usepackage{booktabs}
\usepackage{multirow}
\usepackage{makecell}

\begin{document}

\title{BIG-CBF: Behavior-Imagination-Guided Control Barrier Function with Shared Uncertainty for Mobile Robot Navigation}

\author{
	Shibo~Li$^{1}$,
	Zhongcheng~Wang$^{1}$,
	Jiahe~Cao$^{1}$,
	Jianhua~Yang$^{1,*}$,
	and~Ke~Wu$^{2,*}$%
	\thanks{$^{1}$Shibo Li, Zhongcheng Wang, Jiahe Cao, and Jianhua Yang
		are with the School of Automation, Northwestern Polytechnical University,
		Xi'an, China.
		E-mails: 
		\texttt{lishibo97@mail.nwpu.edu.cn},
		\texttt{wangzhongcheng@mail.nwpu.edu.cn},
		\texttt{chelde@mail.nwpu.edu.cn},
		and \texttt{yangjianhua@nwpu.edu.cn}.}%
	\thanks{$^{2}$Ke Wu is with the Department of Robotics,
		Mohamed bin Zayed University of Artificial Intelligence,
		Abu Dhabi, United Arab Emirates.
		E-mail: \texttt{ke.wu@mbzuai.ac.ae}.}%
	\thanks{$^{*}$Jianhua Yang and Ke Wu are corresponding authors.}%
}

%

\maketitle

\begin{abstract}
	Control barrier functions (CBFs) provide a mathematically grounded framework for enforcing local collision-avoidance constraints in autonomous mobile robots, commonly through optimization-based safety filters. However, a minimum-intervention CBF filter lacks task-level maneuver awareness and may fail to select a productive avoidance direction when multiple distinct maneuvers are locally viable, leading to safe but stalled behavior in geometrically ambiguous environments. This paper presents BIG-CBF, Behavior-Imagination-Guided Control Barrier Function with shared uncertainty, a two-rate navigation architecture that separates low-rate maneuver selection from high-rate safety filtering. Over a short horizon, six closed-loop feedback behaviors are imagined and evaluated using analytic CBF compatibility together with a lightweight objective accounting for task progress, freezing, smoothness, and switching. To reduce planning--execution mismatch, the imagination and execution layers share consistent uncertainty sources for relative-motion delay, obstacle prediction, zero-order-hold motion, and command-execution residuals, while a hard CBF remains the final safety authority. In a 3,600-episode comparative benchmark across nine scenarios, BIG-CBF achieves the highest overall task success rate of 99.78\% while substantially reducing downstream CBF intervention. On a physical omnidirectional robot with onboard Jetson Orin Nano computation, BIG-CBF completes all 15 evaluation runs without a recorded contact event. Matched hardware comparisons against the non-shared variant further show lower CBF intervention energy and activation frequency, supporting improved consistency between maneuver selection and safety-critical execution.
\end{abstract}


\section{Introduction}
Safe local navigation requires a mobile robot to satisfy two coupled objectives: avoiding unsafe states while selecting maneuvers that maintain progress toward the navigation goal. Control barrier functions (CBFs) have emerged as a prominent mechanism for the former, providing modular quadratic-program (QP) safety filters for enforcing forward-invariance conditions \cite{ames2017cbf,hsu2024safetyfilter}.

However, a safety filter determines how a command should be modified to satisfy local safety constraints, rather than which task-progressing avoidance maneuver should be selected. This distinction creates a liveness gap. When a goal-directed command encounters a geometrically symmetric obstacle, the CBF correction may suppress the unsafe approach component without creating a persistent lateral preference. Consequently, the robot may remain safe yet become trapped in a stalled state, demonstrating that forward invariance alone does not imply task-level liveness \cite{mestres2026undesired,chandra2025deadlock}.

To mitigate such local traps, various extensions have been proposed. Backup CBFs \cite{chen2021backup} and Policy-Library CBFs (PL-CBF) \cite{kim2026plcbf} extend safety reasoning beyond the instantaneous command through finite-horizon rollouts of fallback or library policies, while geometric approaches, such as collision-cone CBFs (C3BFs) \cite{tayal2026collision} and turning-circle CBFs \cite{lee2026turning}, encode richer relative-motion or kinematic structure into barrier construction. Concurrently, sampling-based predictive controllers such as Model Predictive Path Integral (MPPI) \cite{williams2017mppi} and safety-integrated variants \cite{yin2023shield,yin2025beyond} combine predictive rollouts with safety mechanisms. Recent works \cite{xiao2025safediffuser,li2026pierflow} have also explored learned generative models for predictive planning and navigation. These predictive and generative approaches provide expressive look-ahead behavior, but typically rely on online trajectory sampling, iterative generation, or offline learned action distributions. For many local avoidance interactions, a compact set of interpretable feedback maneuvers provides a lower-complexity alternative when the dominant decision is the maneuver choice itself.

Practical deployment further complicates this problem. Real-world systems are affected by sampled-data effects, delays, zero-order-hold (ZOH) execution, and model and perception uncertainty \cite{garg2024advances,tan2024zocbf,liu2025sampling}. Recent uncertainty-aware and distributionally robust CBFs explicitly account for uncertain dynamics, perception, or obstacle motion in safety synthesis \cite{li2023uncertainty,long2026distributional,wang2025cvar}. However, consistency between uncertainty-aware safety constraints and an upstream maneuver-selection layer remains less explored, despite the broader importance of cross-layer robustness in layered safety-critical control \cite{compton2025layered}. If candidate maneuvers are evaluated using nominal geometry while the downstream CBF uses uncertainty-inflated geometry, a maneuver judged favorable upstream may repeatedly trigger corrective intervention during execution. This mismatch motivates exposing consistent uncertainty semantics to both maneuver evaluation and command-level safety filtering.

\begin{figure}[tbp]
	\vspace{2mm}
	\centering
	\includegraphics[width=0.95\columnwidth]{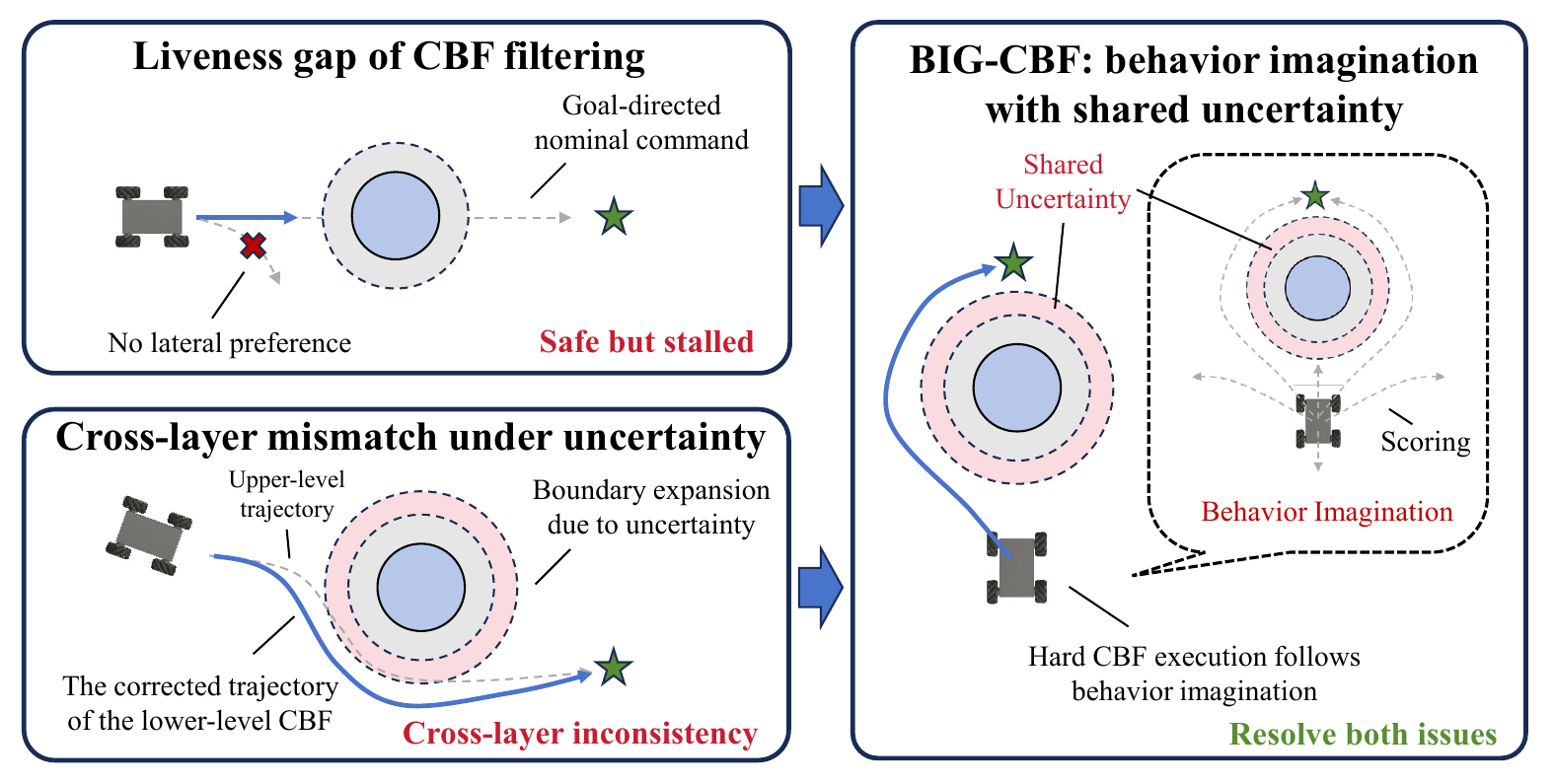} 
	\caption{
		Motivation and core idea of BIG-CBF.  A conventional CBF can preserve local safety yet fail to select a productive avoidance direction, resulting in a safe-but-stalled state in geometrically ambiguous scenes. In layered navigation, a branch that appears feasible under nominal geometry may be repeatedly corrected at execution time when the downstream CBF accounts for uncertainty. BIG-CBF addresses both issues by selecting among a library of closed-loop behaviors through low-rate behavior imagination, while exposing shared uncertainty components to both the imagination and high-rate hard-CBF execution layers.
	}
	\label{fig:motivation}
	\vspace{-2mm}
\end{figure}

Motivated by these observations, we introduce BIG-CBF, Behavior-Imagination-Guided Control Barrier Function with Shared Uncertainty. As illustrated in Fig.~\ref{fig:motivation}, BIG-CBF addresses both maneuver ambiguity and cross-layer uncertainty mismatch through a two-rate architecture. At the low rate, a lightweight imagination layer evaluates a compact library of closed-loop feedback behaviors over a short horizon and selects only a behavior mode. The selected policy then continuously regenerates a state-dependent local subgoal and input command from current measurements at the faster control rate. In parallel, the imagination and execution layers use a common relative-position uncertainty representation constructed from the same uncertainty sources and physical semantics, while allowing layer-dependent calibration. The resulting input command is filtered by a hard CBF, which remains the command-level safety filter.

The primary contributions of this letter are threefold:
\begin{enumerate}
	\item \textbf{Lightweight Behavior Imagination:} We introduce a lightweight maneuver-selection architecture that evaluates a compact library of closed-loop feedback behaviors over a finite horizon. The outer imagination layer selects only a behavior mode, while the selected policy continuously regenerates a state-dependent local subgoal and input command from the latest measurements at the faster execution rate. This design resolves discrete passing, sidestepping, and yielding choices without replaying open-loop imagined trajectories or performing dense trajectory optimization. 
	
	\item \textbf{Cross-Layer Shared-Uncertainty Evaluation:} We formulate a common relative-position uncertainty representation that incorporates delay-induced displacement, obstacle-prediction uncertainty, zero-order-hold motion, and command-execution residuals. The same uncertainty sources and physical semantics are exposed to both behavior-level evaluation and the downstream hard CBF, aligning maneuver selection with the uncertainty-aware constraints encountered during execution without introducing a second safety filter in the imagination layer.
	
	\item \textbf{Extensive Empirical Validation:} We conduct paired evaluations against representative baseline methods over 3,600 simulation episodes across nine scenarios, together with onboard physical-robot experiments. BIG-CBF achieves a 99.78\% overall task success rate in simulation and completes all 15 supervised hardware trials. The evaluation further shows that behavior imagination improves task liveness, while cross-layer uncertainty consistency substantially reduces downstream safety-filter intervention.
\end{enumerate}

\section{Problem Formulation} 
\label{sec:problem} 
\subsection{Robot and Obstacle Models}

\textbf{Robot model.} 
We consider an omnidirectional mobile robot with state $\mathbf{x}=[p_x,p_y,\theta,v_x,v_y,\omega]^\text{T}$, where $\mathbf{p}=[p_x,p_y]^\text{T}$ is the robot position in the planar odometry frame, $\theta$ is the yaw angle, and $(v_x,v_y,\omega)$ are the body-frame velocities. The commanded body-frame velocity is $\mathbf{u}=[u_x,u_y,u_{\omega}]^\text{T}\in\mathbb{R}^{3}$. Throughout this paper, $\mathbf{x}$ and $\mathbf{p}$ denote the robot state and position available to the controller; physical ground-truth quantities are explicitly marked by the superscript ``true''. The realized translational command execution is modeled as 
\begin{equation} 
	\mathbf{v}_{\mathrm{real}} = R(\theta) \left( A\mathbf{u} + \mathbf{d} + \mathbf{r} \right), 
	\label{eq:execution} 
\end{equation} where $\mathbf{v}_{\mathrm{real}}\in\mathbb{R}^{2}$ is the realized velocity in the odometry frame, while $A\mathbf{u}$, $\mathbf{d}$, and $\mathbf{r}$ represent the nominal realized translation, constant bias, and bounded residual in the body frame, respectively. The robot footprint is conservatively approximated by a circle of radius $r_R$.

\textbf{Obstacle model.} 
Let $M$ denote the number of tracked obstacles. At the beginning of each prediction rollout ($k=0$), obstacle $i\in\{1,\ldots,M\}$ is described by its estimated center $\hat{\mathbf{p}}_{i,0}\in\mathbb{R}^{2}$, estimated velocity $\hat{\mathbf{v}}_{i,0}\in\mathbb{R}^{2}$ expressed in the odometry frame, position covariance $\Sigma_{i,0}\in\mathbb{R}^{2\times2}$, and physical radius $r_i$. Over the short imagination horizon, obstacle velocity is assumed constant and its position is propagated as
\begin{equation} 
	\hat{\mathbf{p}}_{i,k+1} = \hat{\mathbf{p}}_{i,k} + \Delta t\,\hat{\mathbf{v}}_{i,k}, 
	\label{eq:obs_prop} 
\end{equation} where $k$ is the discrete prediction index and $\Delta t>0$ is the prediction interval. Prediction uncertainty is approximated by
\begin{equation} 
	\Sigma_{i,k+1} = \Sigma_{i,k} + q_{\Sigma,i}\Delta t I_2, 
	\label{eq:cov_prop} 
\end{equation} where $q_{\Sigma,i}\geq0$ is the obstacle covariance-growth intensity and $I_2$ is the $2\times2$ identity matrix. The predicted obstacle information at step $k$ is collected as
\begin{equation} 
	\mathcal{O}^{k} = \left\{ \hat{\mathbf{p}}_{i,k}, \hat{\mathbf{v}}_{i,k}, \Sigma_{i,k}, r_i \right\}_{i=1}^{M}. 
	\label{eq:predicted_obstacles} 
\end{equation}

\subsection{Navigation Objective and Safety Requirement} 
\textbf{Navigation objective.} 
Given a goal position $\mathbf{p}_g\in\mathbb{R}^{2}$ and goal tolerance $\varepsilon_g>0$, define the goal region as 
\begin{equation} 
	\mathcal{G} = \left\{ \mathbf{x}: \|\mathbf{p}-\mathbf{p}_g\|_2 \leq \varepsilon_g \right\}. 
	\label{eq:goal_region} 
\end{equation} The navigation objective is to reach $\mathcal{G}$ within the task horizon while avoiding persistent stagnation and satisfying the execution-layer collision-avoidance constraints. 

\textbf{Physical safety requirement.} Let $\mathbf{p}^{\mathrm{true}}$ and $\mathbf{p}^{\mathrm{true}}_i$ denote the physical centers of the robot and obstacle $i$, respectively. For a baseline clearance $d_0\geq0$, collision avoidance requires 
\begin{equation} 
	\left\| \mathbf{p}^{\mathrm{true}} - \mathbf{p}^{\mathrm{true}}_i \right\|_2 \geq r_R+r_i+d_0, \qquad \forall i. 
	\label{eq:true_safety_requirement} 
\end{equation}
In practice, the true relative geometry in Eq.~(\ref{eq:true_safety_requirement}) is not directly available due to obstacle-prediction uncertainty, sensing and actuation delay, sampled-data effects, and command-execution errors. BIG-CBF therefore constructs a deterministic uncertainty-aware counterpart of this physical separation requirement and uses the same uncertainty sources for behavior-level evaluation and CBF execution, as developed in Sec.~III.

\section{Proposed Method} 
\label{sec:method} 

\begin{figure*}[t] 
	\vspace*{2mm}
	\centering 
	\includegraphics[width=0.8\textwidth]{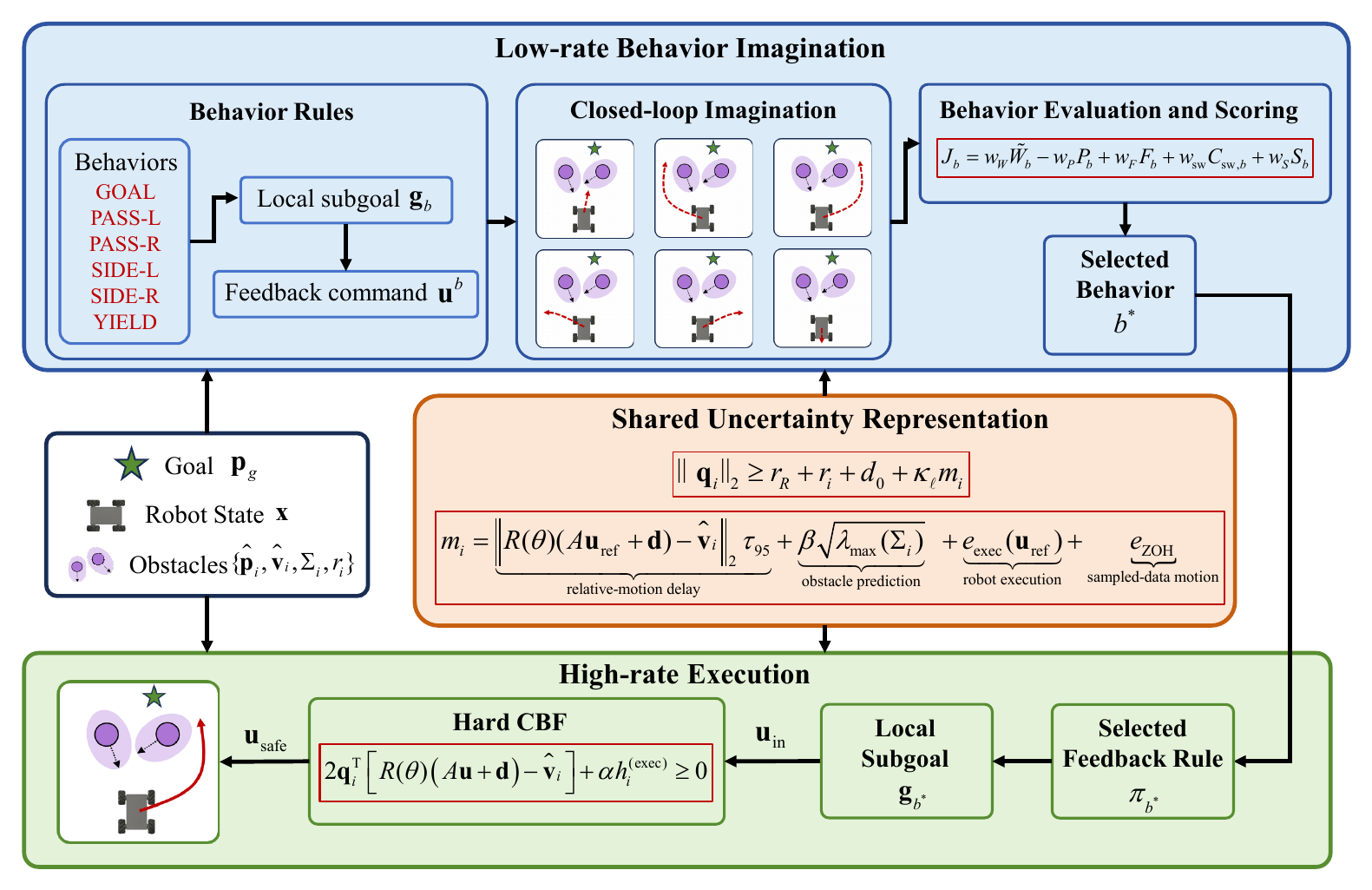} 
	\caption{ 
		Overview of BIG-CBF. The low-rate imagination layer evaluates six closed-loop feedback behaviors and selects only a behavior mode $b^{*}$. At the high-rate execution loop, the selected behavior recomputes a state-dependent local subgoal and $\mathbf{u}_{\mathrm{in}}$ from current measurements, after which the hard CBF generates $\mathbf{u}_{\mathrm{safe}}$. Shared uncertainty components are used for both behavior-level compatibility evaluation and execution-level filtering. 
	} 
	\label{fig:overview}
	\vspace*{-2mm} 
\end{figure*}

BIG-CBF combines three coupled components, as illustrated in Fig.~\ref{fig:overview}. First, a low-rate behavior-imagination layer evaluates a compact set of closed-loop maneuver modes and selects a local strategy. Second, a shared relative-position uncertainty representation aligns behavior evaluation with the uncertainty-aware geometry used during execution. Finally, at the faster control rate, the selected behavior continuously regenerates the input command, which is filtered by a hard CBF before being sent to the robot.

\subsection{Lightweight Behavior Imagination} 
\label{sec:behavior}

BIG-CBF separates low-rate behavior selection from high-rate command execution. Let $T_B$ denote the behavior-selection period and $T_C<T_B$ the control period. At the $n$th behavior update $t_n=nT_B$, the current robot state and obstacle estimates initialize a short-horizon evaluation of
{\small 
\begin{equation} 
	\mathcal{B} = \{ \mathrm{GOAL}, \mathrm{PASS\mbox{-}L}, \mathrm{PASS\mbox{-}R}, \mathrm{SIDE\mbox{-}L}, \mathrm{SIDE\mbox{-}R}, \mathrm{YIELD} \}. 
	\label{eq:behavior_set} 
\end{equation} 
} 

Only the selected mode $b_n^{*}$ is retained; imagined trajectories and commands are not replayed. Between two behavior updates, the selected feedback policy is reevaluated from the latest measurements, 
\begin{equation} 
	\mathbf{u}_{\mathrm{in}}(t) = \pi_{b_n^{*}} \left( \mathbf{x}(t), \mathcal{O}(t), \mathbf{p}_g \right), \qquad t\in[t_n,t_{n+1}), 
	\label{eq:online_behavior_command} 
\end{equation} where $\mathcal{O}(t)$ denotes the current obstacle estimates. Thus, $T_B$ governs maneuver selection, whereas $T_C$ governs feedback command generation and CBF filtering.

\textbf{Feedback behavior rules.} 
Outside the goal region $\mathcal{G}$, define the goal-directed unit vector and its left normal as 
\begin{equation} 
	\mathbf{d}_g = \frac{\mathbf{p}_g-\mathbf{p}} {\|\mathbf{p}_g-\mathbf{p}\|_2}, \qquad \mathbf{n}_g = \begin{bmatrix} -d_{g,y}\\ d_{g,x} \end{bmatrix}. 
	\label{eq:goal_basis} 
\end{equation} 
For $\sigma=+1$ and $\sigma=-1$ denoting left and right, respectively, the behavior-dependent local subgoal is
{\small 
\begin{equation} 
	\mathbf{g}_{b} = \begin{cases} 
		\mathbf{p}_g, & b=\mathrm{GOAL}, \\[1mm] 
		\begin{aligned}[b]
			\mathbf{p} &{}+ 0.85\mathbf{d}_g \\
			&{}+ \sigma\max(d_{\mathrm{pass}},r_R+r_i+0.35)\mathbf{n}_g,
		\end{aligned} & b=\mathrm{PASS}\mbox{-}\sigma, \\[1mm] 
		\mathbf{p} + 0.45\mathbf{d}_g + 0.80\sigma\mathbf{n}_g, & b=\mathrm{SIDE}\mbox{-}\sigma, \\[1mm] 
		\mathbf{p}, & b=\mathrm{YIELD},
	\end{cases} 
	\label{eq:behavior_subgoals} 
\end{equation}
}where $i$ denotes the locally relevant obstacle and $d_{\mathrm{pass}}$ is the minimum lateral passing offset. PASS-L/R retain a goal-directed component while committing to one side; SIDE-L/R emphasize lateral release; and YIELD intentionally suppresses movement when dynamic obstacles are present.

Let $\mathbf{e}_b=\mathbf{g}_b-\mathbf{p}$. A bounded holonomic feedback controller generates the world-frame translational velocity
{\small 
\begin{equation} 
	\mathbf{v}_b = \operatorname{sat}_{v_b^{\max}} \left[ \frac{\mathbf{e}_b}{\|\mathbf{e}_b\|_2} \min \left( v_b^{\max}, (2 {{\text{s}}^{-1}}) \|\mathbf{e}_b\|_2 \right) + (0.35 {{\text{s}}^{-1}})\mathbf{e}_b \right],
	\label{eq:behavior_feedback} 
\end{equation}}where $v_b^{\max}$ is the behavior-dependent speed limit and $\operatorname{sat}_{v_b^{\max}}(\cdot)$ limits the vector norm to $v_b^{\max}$. For YIELD, the command is set to zero. These rules define the feedback policy $\pi_b$ used in Eq.~(\ref{eq:online_behavior_command}).

\textbf{Closed-loop imagination.} For every candidate $b\in\mathcal{B}$, the imagined robot state is initialized as $\hat{\mathbf{x}}_0^{b}=\mathbf{x}(t_n)$. At every prediction step, the command is regenerated from the current imagined state, 
\begin{equation} 
	\mathbf{u}^{b}_k = \pi_b \left( \hat{\mathbf{x}}^{b}_k, \mathcal{O}^{k}, \mathbf{p}_g \right), \qquad k=0,\ldots,N-1. 
	\label{eq:behavior_rollout} 
\end{equation} 
After each step, the robot state is propagated over $\Delta t$ using the execution model in Eq.~(\ref{eq:execution}), while obstacle states and covariances evolve according to Eqs.~(\ref{eq:obs_prop})--(\ref{eq:cov_prop}). Hence both the local subgoal and command are recomputed throughout the rollout.

\textbf{Behavior evaluation and scoring.} Let $\widetilde{W}_b$ denote the accumulated CBF-compatibility penalty constructed from the shared uncertainty model in Sec.~\ref{sec:shared_uncertainty}. The remaining evaluation terms are 
\begin{equation} 
	\begin{aligned} 
		P_b &= \|\mathbf{p}-\mathbf{p}_g\|_2 - \|\hat{\mathbf{p}}^{b}_N-\mathbf{p}_g\|_2, \\ F_b &= \frac{1}{N} \sum_{k=0}^{N-1} \mathbb{I} \left[ \left\| \begin{bmatrix} u^{b}_{x,k}\\ u^{b}_{y,k} \end{bmatrix} \right\|_2 < v_{\mathrm{freeze}} \right], \\ S_b &= \sum_{k=1}^{N-1} \|\mathbf{u}^{b}_k-\mathbf{u}^{b}_{k-1}\|_2^2 + \|\mathbf{u}^{b}_0\|_2^2, \\ C_{\mathrm{sw},b} &= \mathbb{I}[b\neq b_{\mathrm{cur}}], 
	\end{aligned} 
	\label{eq:behavior_terms} 
\end{equation} 
where $\hat{\mathbf{p}}^{b}_N$ is the position component of the terminal imagined state, $v_{\mathrm{freeze}}$ is the translational freezing threshold, and $b_{\mathrm{cur}}$ is the currently retained mode.

Candidates passing the robust barrier test defined in Sec.~\ref{sec:shared_uncertainty} are ranked by
\begin{equation} 
	J_b = w_W\widetilde{W}_b - w_P P_b + w_F F_b + w_{\mathrm{sw}}C_{\mathrm{sw},b} + w_S S_b, 
	\label{eq:behavior_cost} 
\end{equation}
with positive weights $w_W,w_P,w_F,w_{\mathrm{sw}},w_S$. The lowest-cost accepted candidate is selected subject to a minimum dwell time and a relative switching-improvement threshold. If all regular behaviors fail the acceptance test, the controller invokes the predefined reverse-release fallback. The numerical parameters are reported in Sec.~IV.

\subsection{Cross-Layer Shared-Uncertainty Evaluation} 
\label{sec:shared_uncertainty}

Shared uncertainty aligns behavior-level evaluation with the uncertainty-aware geometry encountered during execution without introducing a second safety filter in the imagination layer.

\textbf{Relative-position uncertainty representation.} 
We aggregate their effect into the relative-position error $\boldsymbol{\epsilon}^{\mathrm{rel}}_i$. Define
\begin{equation} 
	\mathbf{q}_i = \mathbf{p}-\hat{\mathbf{p}}_i, \qquad \boldsymbol{\epsilon}^{\mathrm{rel}}_i = (\mathbf{p}^{\mathrm{true}}-\mathbf{p}) - (\mathbf{p}^{\mathrm{true}}_i-\hat{\mathbf{p}}_i). 
	\label{eq:relative_position_error} 
\end{equation}
The true relative position is therefore
\begin{equation} 
	\mathbf{p}^{\mathrm{true}} - \mathbf{p}^{\mathrm{true}}_i = \mathbf{q}_i + \boldsymbol{\epsilon}^{\mathrm{rel}}_i.
\end{equation}

For layer $\ell\in\{\mathrm{img},\mathrm{exec}\}$, let $\kappa_\ell m_i$ denote the adopted relative-position uncertainty envelope,
\begin{equation} 
	\|\boldsymbol{\epsilon}^{\mathrm{rel}}_i\|_2 \leq \kappa_\ell m_i, 
	\label{eq:relative_error_bound} 
\end{equation}
where $m_i$ is the common uncertainty radius and $\kappa_\ell$ allows layer-dependent calibration.

Applying the reverse triangle inequality gives
\begin{equation} 
	\begin{aligned} \left\| \mathbf{p}^{\mathrm{true}} - \mathbf{p}^{\mathrm{true}}_i \right\|_2 &= \left\| \mathbf{q}_i + \boldsymbol{\epsilon}^{\mathrm{rel}}_i \right\|_2 \\ &\geq \|\mathbf{q}_i\|_2 - \|\boldsymbol{\epsilon}^{\mathrm{rel}}_i\|_2 \\ &\geq \|\mathbf{q}_i\|_2 - \kappa_\ell m_i . \end{aligned}
\end{equation}
Therefore, a sufficient instantaneous condition for satisfying
Eq.~(\ref{eq:true_safety_requirement}) is
\begin{equation} 
	\|\mathbf{q}_i\|_2 \geq r_R+r_i+d_0+\kappa_\ell m_i. 
	\label{eq:robust_distance_condition} 
\end{equation}
Accordingly, define
\begin{equation} 
	\begin{aligned} 
		D_i^{(\ell)} = r_R+r_i+d_0+\kappa_\ell m_i, \\
		\qquad h_i^{(\ell)} = \|\mathbf{q}_i\|_2^2 - \left(D_i^{(\ell)}\right)^2.
	\end{aligned}
	\label{eq:shared_barrier} 
\end{equation}
Thus, $D_i^{(\ell)}$ follows directly from a bound on the uncertain relative position. For the execution layer, the resulting robustness statement is conditional on the adopted bound in Eq.~(\ref{eq:relative_error_bound}) containing the actual relative-position error.

\textbf{Cross-layer uncertainty sharing.}
Let $\mathbf{u}_{\mathrm{ref}}$ denote the command associated with the layer being evaluated: $\mathbf{u}_{\mathrm{ref}}=\mathbf{u}^{b}_k$ during imagination and $\mathbf{u}_{\mathrm{ref}}=\mathbf{u}_{\mathrm{in}}$ during execution. The common relative-position uncertainty radius is
\begin{equation} 
	\begin{aligned} 
		m_i ={}& \underbrace{ \left\| R(\theta) (A\mathbf{u}_{\mathrm{ref}}+\mathbf{d}) - \hat{\mathbf{v}}_i \right\|_2\tau_{95} }_{\text{relative-motion delay}} + \underbrace{ \beta\sqrt{\lambda_{\max}(\Sigma_i)} }_{\text{obstacle prediction}} \\ &+ \underbrace{ e_{\mathrm{exec}}(\mathbf{u}_{\mathrm{ref}}) }_{\text{robot execution}} + \underbrace{ e_{\mathrm{ZOH}} }_{\text{sampled-data motion}} . 
	\end{aligned} 
	\label{eq:uncertainty_margin} 
\end{equation}
Here $\tau_{95}$ is the adopted perception-to-actuation delay, $\beta\sqrt{\lambda_{\max}(\Sigma_i)}$ is the covariance-based obstacle prediction radius, $e_{\mathrm{exec}}$ bounds displacement induced by execution residuals, and $e_{\mathrm{ZOH}}$ accounts for inter-sample zero-order-hold motion. Eq.~(\ref{eq:uncertainty_margin}) is reevaluated with the corresponding predicted quantities during imagination. In particular, for candidate $b$ at step $k$, $h_{i,k}^{(\mathrm{img}),b}$ denotes Eq.~(\ref{eq:shared_barrier}) evaluated with $\ell=\mathrm{img}$, the imagined robot state $\hat{\mathbf{x}}_k^b$, obstacle prediction $(\hat{\mathbf{p}}_{i,k},\hat{\mathbf{v}}_{i,k},\Sigma_{i,k})$, and $\mathbf{u}_{\mathrm{ref}}=\mathbf{u}_k^b$.

A candidate is rejected when
\begin{equation} 
	\min_{k,i} h_{i,k}^{(\mathrm{img}),b} < -\varepsilon_h, 
	\label{eq:behavior_acceptance} 
\end{equation}
where $\varepsilon_h\geq0$ is the barrier-acceptance tolerance. For an accepted candidate, CBF compatibility is evaluated analytically without solving a future QP. Let $\alpha>0$ be the CBF gain and define the scalar CBF margin along the imagined rollout as
{\small 
\begin{equation} 
	\gamma_{i,k}^{b}(\mathbf{u}) = 2 (\hat{\mathbf{p}}_k^b-\hat{\mathbf{p}}_{i,k})^\text{T} \left[ R(\hat{\theta}_k^b) (A\mathbf{u}+\mathbf{d}) - \hat{\mathbf{v}}_{i,k} \right] + \alpha h_{i,k}^{(\mathrm{img}),b}, 
\end{equation}
}where $\hat{\mathbf{p}}_k^b$ and $\hat{\theta}_k^b$ are the position and yaw components of $\hat{\mathbf{x}}_k^b$. Since $\gamma_{i,k}^{b}(\mathbf{u})$ is affine in $\mathbf{u}$, the accumulated normalized violation is
\begin{equation} 
	\widetilde{W}_b = \sum_{k=0}^{N-1} \sum_{i=1}^{M} \frac{ \left[ \max \left( -\gamma_{i,k}^{b}(\mathbf{u}_k^b), 0 \right) \right]^2 }{ \| \nabla_{\mathbf{u}} \gamma_{i,k}^{b} \|_2^2+\epsilon }, 
	\label{eq:analytic_workload} 
\end{equation}
where $\epsilon>0$ is a small numerical regularizer. With multiple coupled constraints and actuator bounds, $\widetilde W_b$ is used only as an analytic intervention surrogate, and no CBF-QP is solved inside the imagination layer.

Hence, the two layers share the same uncertainty sources and physical semantics, while $\kappa_\ell$ allows layer-specific calibration. Behavior imagination uses the resulting geometry only for candidate evaluation, whereas the execution layer uses $h_i^{(\mathrm{exec})}$ in the actual hard safety filter.

\subsection{Hard CBF Execution Filter} 
\label{sec:hard_cbf}

\textbf{CBF constraint.} 
At each control update, $D_i^{(\mathrm{exec})}$ is evaluated from the current uncertainty profile and $\mathbf{u}_{\mathrm{in}}$, and is held fixed during the current optimization. The execution filter then
enforces
\begin{equation} 
	2\mathbf{q}_i^\text{T} \left[ R(\theta) (A\mathbf{u}+\mathbf{d}) - \hat{\mathbf{v}}_i \right] + \alpha h_i^{(\mathrm{exec})} \geq 0. 
	\label{eq:cbf_condition} 
\end{equation}
With $D_i^{(\mathrm{exec})}$ fixed within the current solve, Eq.~(\ref{eq:cbf_condition}) is affine in $\mathbf{u}$. This constraint is used as a sampled command-level safety filter.

\textbf{Minimum-intervention filtering.} 
The final safety filter computes
\begin{equation} 
	\begin{aligned} 
		\mathbf{u}_{\mathrm{safe}} = \arg\min_{\mathbf{u}} \quad& (\mathbf{u}-\mathbf{u}_{\mathrm{in}})^\text{T} H (\mathbf{u}-\mathbf{u}_{\mathrm{in}}) \\ \mathrm{s.t.}\quad& \text{Eq.~(\ref{eq:cbf_condition})},\quad \forall i, \\ & \mathbf{u}_{\min} \leq \mathbf{u} \leq \mathbf{u}_{\max}, 
	\end{aligned} 
	\label{eq:hard_cbf_qp} 
\end{equation}
where $H\succ0$ weights modification of the command channels and $\mathbf{u}_{\min},\mathbf{u}_{\max}$ are the actuator bounds. If $\mathbf{u}_{\mathrm{in}}$ already satisfies all constraints, the minimum-intervention solution is $\mathbf{u}_{\mathrm{safe}}=\mathbf{u}_{\mathrm{in}}$; otherwise, the CBF modifies it within the feasible set. 

\section{Simulation Results} 
\label{sec:simulation} 
\subsection{Simulation Setup} 
\label{sec:sim_setup} 

\begin{figure}[htbp] 
	\vspace{2mm}
	\centering 
	\includegraphics[width=0.95\columnwidth]{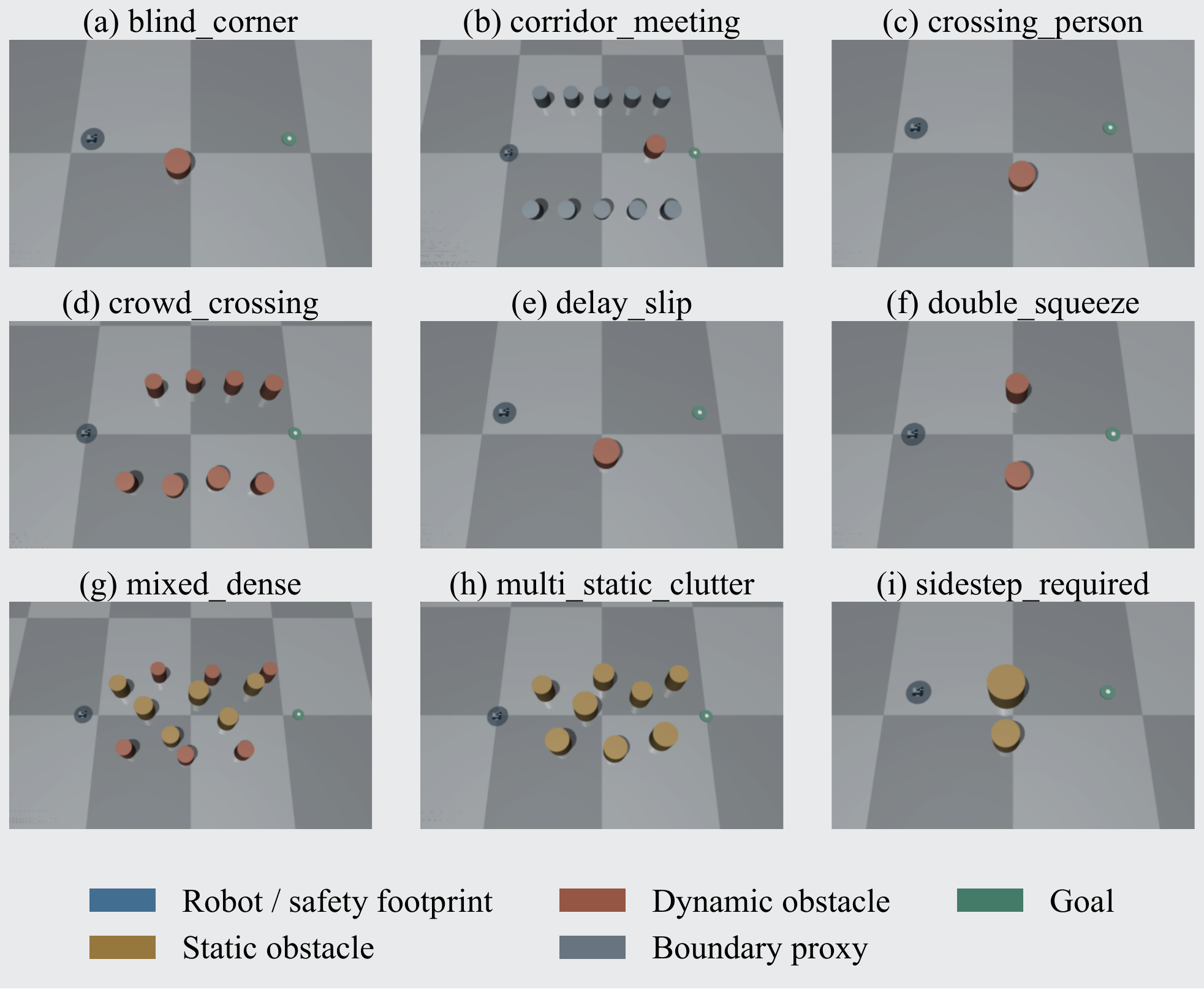} 
	\caption{ 
		Nine simulation benchmark scenarios. (a) blind-corner emergence, (b) corridor meeting, (c) single-person crossing, (d) crowd crossing, (e) delay and slip, (f) double squeeze, (g) mixed dense traffic, (h) multi-static clutter, and (i) required sidestepping.
	} 
	\label{fig:simulation_scenarios} 
	\vspace{-2mm}
\end{figure}

We evaluate BIG-CBF in an omnidirectional-robot simulator that explicitly models command latency, affine execution scaling, bounded execution residuals, and lateral slip. As illustrated in Fig.~\ref{fig:simulation_scenarios}, nine scenarios cover limited visibility, narrow-space negotiation, dynamic interactions, execution uncertainty, dense clutter, and mandatory lateral maneuvering. All methods use identical observations, actuator limits, execution disturbances, and the same final hard CBF filter. We compare CBF-only \cite{ames2017cbf}, Backup-CBF+CBF \cite{chen2021backup}, PL-CBF+CBF \cite{kim2026plcbf}, APF+CBF \cite{khatib1986apf}, DWA+CBF \cite{fox1997dwa}, MPPI+CBF \cite{williams2017mppi}, BIG-CBF-NoShared, and BIG-CBF. CBF-only directly provides goal-tracking commands to the common CBF, whereas Backup-CBF and PL-CBF act as policy shields before the same execution filter. APF and DWA operate at the control rate. DWA uses a 1.5~s rolling horizon, matching the prediction horizon of BIG-CBF. MPPI is configured with the same 2~Hz outer-loop frequency and 1.5~s prediction horizon as BIG-CBF, and MPPI itself uses 512 sampled trajectories and two sampling updates per planning cycle. Each method--scenario combination is evaluated using the same 50 random seeds, resulting in $8\times9\times50=3600$ formal episodes.

\textbf{Parameters.} The simulation instantiates the symbols introduced in Secs.~II--III as summarized in Table~\ref{tab:sim_parameters}. The behavior-selection and execution periods are $T_B=0.5$ s and $T_C=0.05$ s, respectively. Behavior imagination uses $N=15$ prediction steps with $\Delta t=0.1$ s. The latency and execution-residual terms were obtained from an independent calibration sequence.

\begin{table}[htbp] 
	\centering 
	\caption{Principal Simulation Parameters} 
	\label{tab:sim_parameters} 
	\scriptsize 
	\setlength{\tabcolsep}{3.5pt} 
	\begin{tabular}{lcc} 
		\toprule 
		\textbf{Quantity} & \textbf{Symbol} & \textbf{Value} \\ 
		\midrule 
		Behavior-selection period & $T_B$ & 0.50 s (2 Hz) \\ 
		Execution / CBF period & $T_C$ & 0.05 s (20 Hz) \\ 
		Perception rate & -- & $\approx10$ Hz \\ 
		Imagination horizon & $N,\Delta t$ & $15,\ 0.10$ s \\ 
		Goal tolerance & $\varepsilon_g$ & 0.15 m \\ 
		Freeze threshold & $v_{\mathrm{freeze}}$ & 0.05 m/s \\ 
		Barrier acceptance tolerance & $\varepsilon_h$ & 0.10 $\mathrm{m}^2$ \\ 
		Base clearance & $d_0$ & 0.10 m \\ 
		Delay parameter & $\tau_{95}$ & 0.10 s \\ 
		Covariance scale & $\beta$ & 2.0 \\ 
		ZOH displacement & $e_{\mathrm{ZOH}}$ & 0.025 m \\ 
		Layer scaling & $\kappa_{\mathrm{img}},\kappa_{\mathrm{exec}}$ & $1.0,\ 1.0$ \\ 
		Score weights & \makecell{$(w_W,w_P,w_F,$ \\ $w_{\mathrm{sw}},w_S)$} & \makecell{$(1,1.5,3.0,$ \\ $0.5,0.05)$} \\
		\bottomrule 
	\end{tabular} 
\end{table}

\textbf{Evaluation metrics.} 
Task success denotes reaching the goal region $\mathcal{G}$ before the scenario timeout. In addition to completion time and minimum geometric clearance, we measure the burden placed on the common CBF using 
\begin{equation} 
	E_{\mathrm{CBF}} = \sum_{k=0}^{K-1} (\mathbf{u}_{\mathrm{safe},k}-\mathbf{u}_{\mathrm{in},k})^\text{T} R_u (\mathbf{u}_{\mathrm{safe},k}-\mathbf{u}_{\mathrm{in},k}), 
	\label{eq:sim_cbf_energy} 
\end{equation} where $R_u$ is fixed across all methods. This is a weighted command-correction metric. Lower values indicate that the upstream controller more frequently generates commands already compatible with the common CBF. We also report the common-CBF activation ratio, defined as the fraction of control steps for which at least one CBF constraint is active, and the hard-QP infeasibility rate.

\subsection{Comparative Analysis} 
\label{sec:baseline_sim} 

\textbf{Aggregate performance.} 
Table~\ref{tab:overall_sim} reports aggregate performance over the 450 episodes of each method. All eight methods recorded zero simulated collisions under the common hard CBF execution layer; the main differences therefore concern task liveness and the amount of downstream CBF intervention. BIG-CBF completed 449/450 episodes, corresponding to 99.78\% task success. More importantly, it required substantially less correction from the common safety filter: mean $E_{\mathrm{CBF}}$ was 0.02, compared with 0.83 for DWA+CBF and 0.45 for MPPI+CBF, while the CBF activation ratio decreased to 9.10\% from 30.24\% and 36.94\%, respectively. These results indicate that behavior imagination more frequently produces $\mathbf{u}_{\mathrm{in}}$ inside or near the downstream CBF-admissible set. BIG-CBF also achieved the largest mean minimum clearance, 0.55 m.

\begin{table}[htbp]
	\vspace*{2mm}
	\caption{Aggregated Performance Across 9 Scenarios}
	\centering
	\label{tab:overall_sim}
	\scriptsize
	\setlength{\tabcolsep}{2.8pt} 
		\begin{tabular}{lcccccc}
			\toprule
			\textbf{Method} & \textbf{Success} & \textbf{Time} & \textbf{Clearance} & \textbf{$\boldsymbol{E_{\mathrm{CBF}}}$} & \textbf{Active} & \textbf{QP Infeas.} \\
			& (\%) & (s) & (m) &   & (\%) & (\%) \\
			\midrule
			CBF-only & 68.67 & 15.13 & 0.35 & 7.03 & 62.51 & 0.04 \\
			Backup-CBF + CBF & 76.00 & 13.70 & 0.35 & 4.97 & 39.06 & 0.06 \\
			PL-CBF + CBF & 74.89 & 14.51 & 0.33 & 6.08 & 44.29 & 0.08 \\
			APF + CBF & 82.44 & 14.52 & 0.51 & 0.17 & 14.73 & 0.00 \\
			DWA + CBF & 93.78 & 11.31 & 0.41 & 0.83 & 30.24 & 0.03 \\
			MPPI + CBF & 83.33 & 12.35 & 0.41 & 0.45 & 36.94 & 0.03 \\
			BIG-CBF-NoShared & 98.89 & \textbf{11.21} & 0.52 & 0.10 & 15.64 & 0.00 \\
			\textbf{BIG-CBF} & \textbf{99.78} & 11.72 & \textbf{0.55} & \textbf{0.02} & \textbf{9.10} & \textbf{0.00} \\
			\bottomrule
		\end{tabular}
\end{table}

\begin{figure}[t] 
	\vspace{2mm}
	\centering 
	\includegraphics[width=0.94\columnwidth]{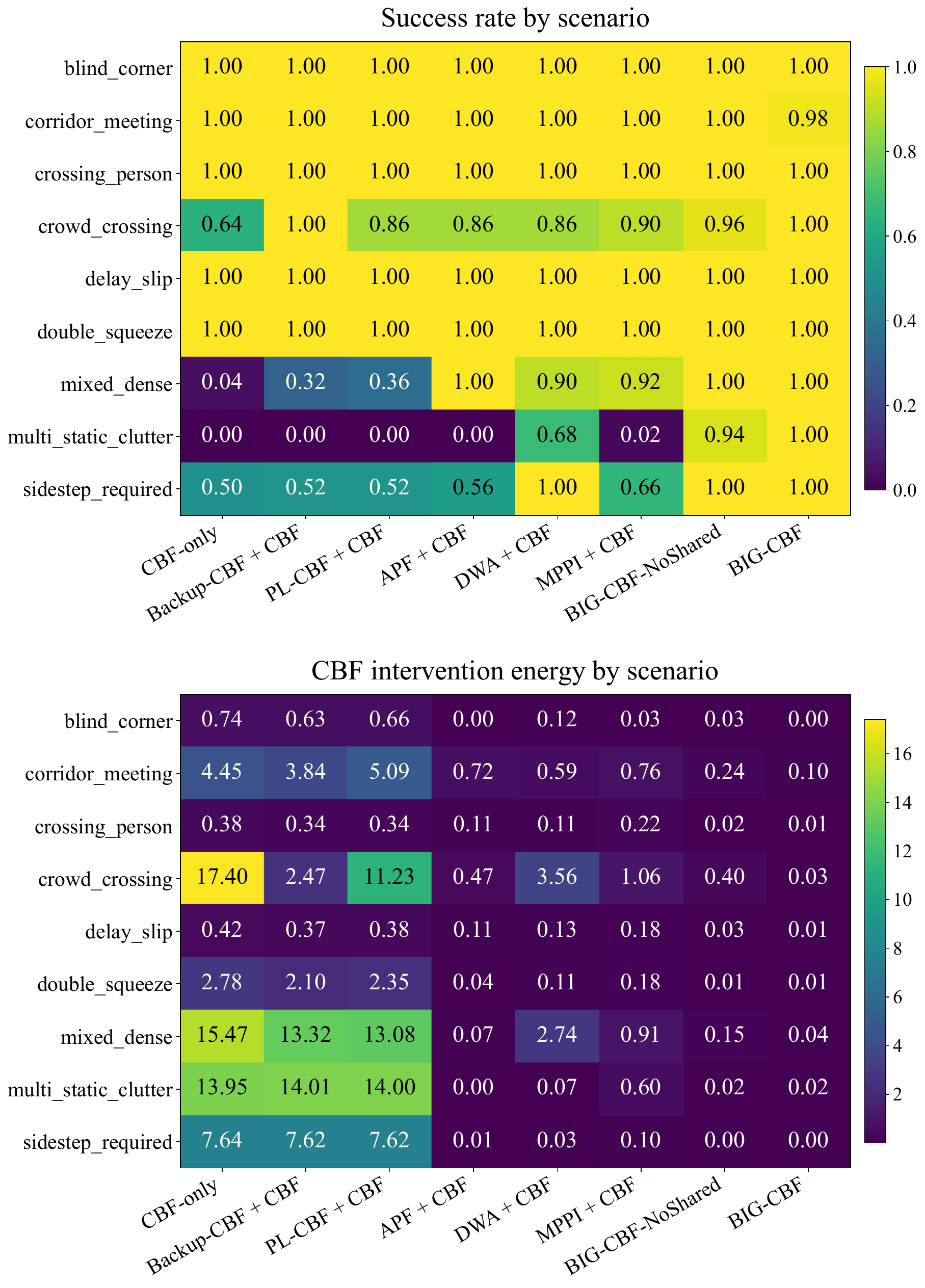} 
	\caption{ 
		Scenario-level performance across the nine simulation environments. The largest success-rate differences occur in topologically ambiguous scenarios such as required sidestepping and multi-static clutter, while several simpler scenarios saturate at 100\%. All methods use the same final hard CBF execution filter. 
		} 
	\label{fig:scenario_heatmaps} 
	\vspace{-2mm}
\end{figure}

\textbf{Scenario-level liveness.} 
Fig.~\ref{fig:scenario_heatmaps} shows that the aggregate difference is concentrated in scenarios requiring persistent maneuver selection. In required sidestepping, CBF-only, Backup-CBF, PL-CBF, and APF achieved 50\%, 52\%, 52\%, and 56\% success, respectively; DWA+CBF and both BIG-CBF variants achieved 100\%, while matched-horizon MPPI+CBF achieved 66\%. Multi-static clutter was more discriminative: CBF-only, Backup-CBF, PL-CBF, and APF all achieved 0\%, DWA achieved 68\%, MPPI 2\%, BIG-CBF-NoShared 94\%, and BIG-CBF 100\%. These cases support the intended role of behavior imagination in maintaining a productive passing or sidestepping mode rather than relying on the pointwise CBF to create such a preference. Conversely, BIG-CBF achieved 98\% in corridor meeting, where all comparison methods achieved 100\%; its single failure was a collision-free timeout, illustrating that persistent behavior commitment and uncertainty inflation can occasionally reduce efficiency in narrow but feasible geometries.

\textbf{Computation.} 
BIG-CBF required 55.68 ms P95 for its $T_B=0.5$ s imagination cycle, corresponding to 11.14\% of the available outer-loop period. Matched MPPI required 77.80 ms (15.56\%), whereas DWA required 2.12 ms P95 at its 20~Hz replanning rate, or 4.24\% of its 50-ms period. The common CBF required approximately 0.80 ms P95 for BIG-CBF. No evaluated outer planner missed its scheduling deadline.

\subsection{Shared-Uncertainty Ablation} 
\label{sec:shared_ablation}

\textbf{Ablation design.} 
BIG-CBF-NoShared isolates the contribution of cross-layer uncertainty consistency. It retains the same behavior library, feedback policies, horizon, switching logic, and final hard CBF as BIG-CBF, but replaces the imagination-layer uncertainty construction in Eq.~(\ref{eq:uncertainty_margin}) with a fixed 0.30-m geometric margin. Thus, any difference between the two variants arises from how candidate behaviors are evaluated upstream.

\textbf{Safety-filter burden.} 
BIG-CBF-NoShared achieved 98.89\% success compared with 99.78\% for BIG-CBF. More clearly, uncertainty sharing reduced the common-CBF activation ratio from 15.64\% to 9.10\% and reduced mean $E_{\mathrm{CBF}}$ from 0.10 to 0.02. Mean minimum clearance increased from 0.52 m to 0.55 m. These results are consistent with the intended effect of shared uncertainty: candidate maneuvers are evaluated under uncertainty semantics consistent with those encountered by the downstream CBF, reducing subsequent command correction.

\textbf{Conservatism trade-off.} 
The improved cross-layer consistency incurs a modest efficiency cost. BIG-CBF required a mean completion time of 11.72 s, compared with 11.21 s for BIG-CBF-NoShared. The additional stopped time was associated with deliberately selected YIELD behavior rather than persistent unintended stagnation. Shared uncertainty should therefore be interpreted as trading modest proactive conservatism for lower downstream intervention and larger simulated separation.

\section{Real-World Experiments}
\label{sec:realworld}
\subsection{Platform, Calibration, and Implementation} 
\label{sec:real_platform}
\textbf{Platform and perception.} 
Physical experiments were conducted on a Rosmaster X3 Mecanum-wheeled robot equipped with an NVIDIA Jetson Orin Nano (4 GB RAM) running ROS~2 Humble. Robot state estimation combined wheel odometry and IMU measurements, while LiDAR-based obstacle perception operated at approximately 10 Hz. Detected obstacles were clustered and associated into at most eight tracked entities using constant-velocity Kalman filters with static/dynamic classification hysteresis. All perception, behavior imagination, CBF filtering, and command generation were executed onboard. The simulation timing configuration was retained, with $T_C=0.05$ s for the execution/CBF loop and $T_B=0.5$ s for BIG-CBF behavior imagination. The translational speed limit was 0.8 m/s, while SIDE-L/R used a lateral speed limit of 0.55 m/s.

\begin{figure*}[tbp] 
	\vspace*{2mm}
	\centering
	\includegraphics[width=0.8\textwidth]{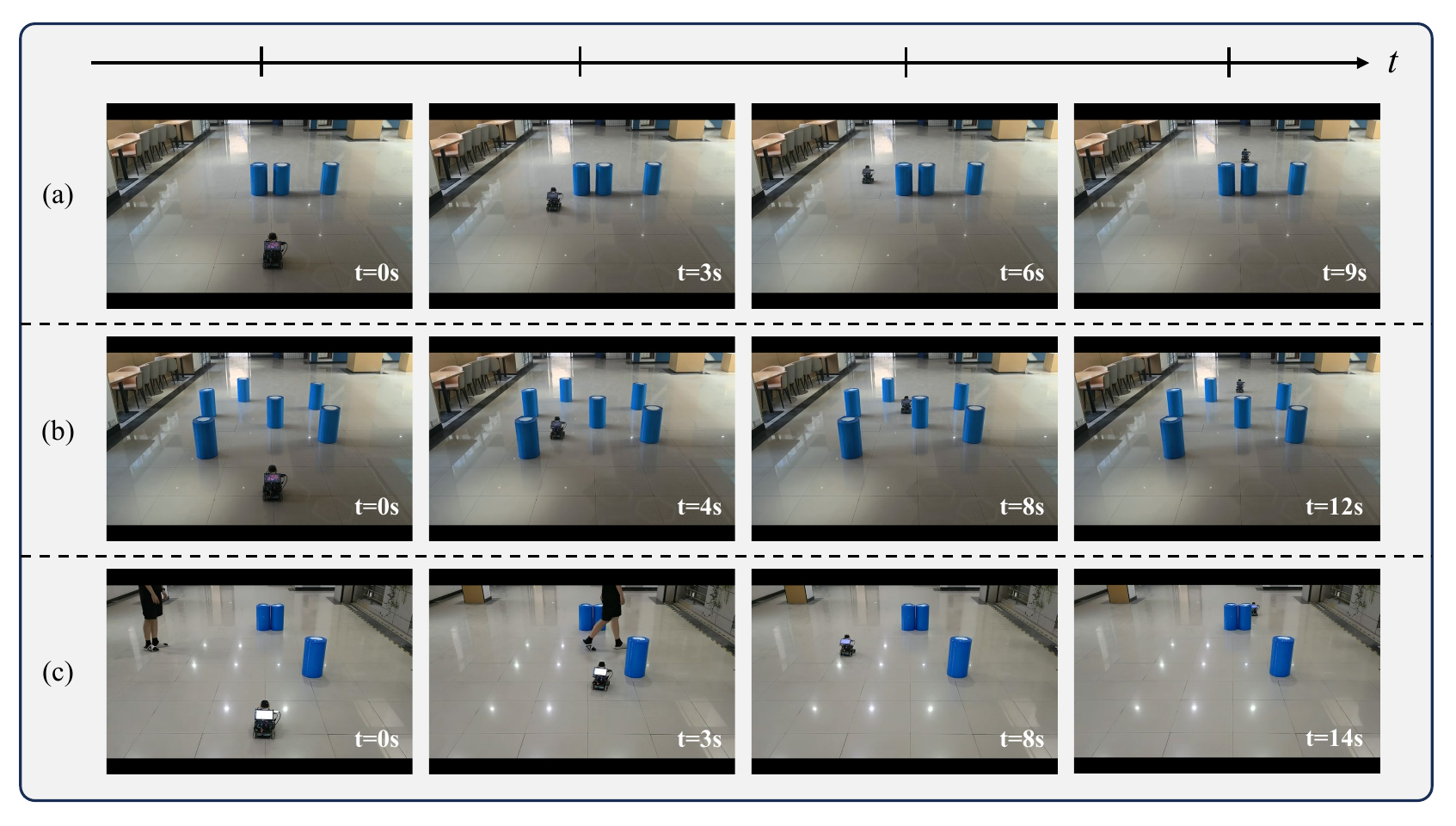} 
	\caption{Representative real-world executions of BIG-CBF in three navigation scenarios: (a) required sidestepping, (b) static clutter, and (c) pedestrian crossing. For each scenario, four snapshots are shown from left to right in temporal order. The sequences illustrate how BIG-CBF maintains a persistent local maneuver while continuously updating the executed command from current onboard perception. Full temporal executions are provided in the supplementary video. } 
	\label{fig:real_world} 
	\vspace*{-2mm}
\end{figure*}

\textbf{Physical uncertainty profile.} 
The uncertainty model was calibrated separately from the synthetic simulation profile. The robot footprint was conservatively represented by a circle with $r_R=0.15$ m. The physical profile used $d_0=0.05$ m, $\beta=1$, and $e_{\mathrm{ZOH}}=0.0716$ m, together with latency and command-dependent execution-residual terms obtained from hardware measurements. The execution layer used the full profile ($\kappa_{\mathrm{exec}}=1$); in dynamic scenes, the imagination layer used $\kappa_{\mathrm{img}}=0.8$ for the non-base uncertainty terms. BIG-CBF-NoShared retained the same execution CBF but replaced the imagination-side shared-uncertainty model with a fixed 0.30 m margin.

\subsection{Hardware Evaluation Results} 
\label{sec:real_results} 

\textbf{Evaluation protocol.} 
The eight methods were evaluated in three representative physical navigation scenarios: required sidestepping, static clutter, and pedestrian crossing. Each method--scenario combination was repeated five times, resulting in 120 supervised trials. Fig.~\ref{fig:real_world} qualitatively illustrates BIG-CBF behavior in the three test environments.

\begin{table}[htbp]
	\caption{Overall performance in 3 real-world scenarios}
	\label{tab:real_results}
	\centering
	\resizebox{\columnwidth}{!}{
		\begin{tabular}{lcccccc}
			\toprule
			\textbf{Method} & \textbf{Success} & \textbf{Time} & \textbf{Clearance} & \textbf{$\boldsymbol{E_{\mathrm{CBF}}}$} & \textbf{Active} & \textbf{QP Infeas.} \\
			&  & (s) & (m) &   & (\%) & (\%) \\
			\midrule
			CBF-only & 7/15 & 21.62 & \textbf{0.29} & 11.04 & 88.60 & 0.12 \\
			Backup-CBF + CBF & 8/15 & 19.80 & 0.21 & 1.65 & 59.60 & 0.09 \\
			PL-CBF + CBF & 9/15 & 19.92 & 0.16 & 1.49 & 55.20 & 0.43 \\
			APF + CBF & 1/15 & 29.75 & \textbf{0.29} & 0.28 & 29.60 & 0.04 \\
			DWA + CBF & 13/15 & 12.32 & 0.14 & 0.88 & 47.30 & 1.42 \\
			MPPI + CBF & 9/15 & 20.56 & 0.27 & 0.40 & 42.85 & 0.10 \\
			BIG-CBF-NoShared & \textbf{15/15} & \textbf{11.79} & 0.24 & 0.60 & 39.30 & 0.18 \\
			\textbf{BIG-CBF} & \textbf{15/15} & 13.18 & \textbf{0.29} & \textbf{0.26} & \textbf{26.9} & \textbf{0.00} \\
			\bottomrule
		\end{tabular}
	}
\end{table}

\textbf{Task performance.} 
BIG-CBF completed all 15 physical trials, as did BIG-CBF-NoShared, while DWA+CBF completed 13/15, MPPI+CBF 9/15, and APF+CBF 1/15. Together with the simulation results, these observations support the role of persistent behavior selection in local geometries where purely reactive commands may fail to maintain task progress.

\textbf{Shared-uncertainty effect.} 
The matched comparison between BIG-CBF and BIG-CBF-NoShared provides more direct evidence for the effect of cross-layer uncertainty consistency. Shared uncertainty reduced common-CBF intervention energy by 0.34 and reduced the CBF activation rate by 12.40 percentage points. This reduction in downstream intervention was accompanied by an increase in mean completion time from 11.79 s to 13.18 s, indicating a modest conservatism--efficiency trade-off.

\textbf{Onboard computation.} 
BIG-CBF remained within its onboard behavior-imagination budget on the Jetson Orin Nano, with a planner P95 latency of 141.0 ms, corresponding to 28.20\% of the available $T_B=0.5$ s period. The $T_C=0.05$ s exact active-set CBF achieved a P95 solve time of 3.86 ms and recorded a 0.00\% QP failure rate. These measurements demonstrate that the two-rate architecture can be executed fully onboard under the computational constraints of the tested platform.

\section{Conclusion} 
\label{sec:conclusion}
This paper presented BIG-CBF, a two-rate mobile-robot navigation framework that combines behavior-level maneuver selection with hard-CBF-based safety filtering under a shared uncertainty representation. Across 3,600 simulation episodes, BIG-CBF achieved a 99.78\% task success rate while substantially reducing downstream CBF intervention, and the ablation results further showed that cross-layer uncertainty consistency lowers safety-filter burden while maintaining conservative separation. In physical experiments, BIG-CBF completed all 15 supervised trials and demonstrated onboard computational feasibility on a Mecanum robot equipped with a Jetson Orin Nano. These results demonstrate that lightweight behavior imagination can improve task liveness while remaining compatible with high-rate CBF execution. Future work will extend the framework to richer geometric representations, more principled uncertainty models, and adaptive behavior generation in complex dynamic environments.




%

\bibliographystyle{IEEEtran}
\bibliography{referencecbf}

%
%
%
%
%
%
%
%
%

\end{document}